%% file: ijcai26.tex
\documentclass{article}
\usepackage{ijcai26}

\usepackage{times}
\usepackage{soul}
\usepackage{url}
\usepackage[hidelinks]{hyperref}
\usepackage[utf8]{inputenc}
\usepackage[small]{caption}
\usepackage{graphicx}
\usepackage{amsmath}
\usepackage{amsthm}
\usepackage{booktabs}
\usepackage{algorithm}
\usepackage{algorithmic}
\usepackage[switch]{lineno}
\usepackage{graphicx}
\usepackage{subcaption}
\usepackage{xcolor}
\usepackage{amsmath}
\usepackage{amssymb}
\usepackage{comment}

\usepackage{booktabs}   
\usepackage{multirow}   
\usepackage[table]{xcolor} 
\usepackage{graphicx}   
\definecolor{myred}{RGB}{204,36,29}

\title{RegionCache: Semantic-Aware Region Reuse for Efficient Multi-Turn Image Generation}

\author{
    Author Name
    \affiliations
    Affiliation
    \emails
    email@example.com
}

\author{
Peizheng Li\and
Xin Ai\and
Hanyuan Liu\and
Qiange Wang\thanks{Corresponding authors.}\and
Yanfeng Zhang\footnotemark[1]\\
\affiliations
School of Computer Science and Engineering, Northeastern University, China\\
\emails
\{lipz2, aix, liuhanyuan\}@mails.neu.edu.cn,
\{wangqg, zhangyf\}@mail.neu.edu.cn
}

\begin{document}

\maketitle

\begin{abstract}

Real-world image generation often involves multi-turn editing, where users iteratively modify small regions while most image content remains unchanged. However, existing diffusion transformer (DiT)-based editing pipelines recompute the entire image at every turn, causing substantial redundant computation. Existing DiT acceleration methods further ignore semantic correspondence across prompts, leading to unnecessary recomputation or unsafe reuse that harms editing quality. To address this, we propose \textbf{RegionCache}, a semantic-aware reuse framework for multi-turn image editing that selectively reuses diffusion states from unchanged regions. RegionCache detects reusable regions through semantic overlap between consecutive prompts and cross-attention localization, and adopts an adaptive reuse schedule based on prompt similarity and contextual consistency. Experiments on PixArt-$\alpha$ demonstrate that RegionCache achieves $1.43\times$--$2.55\times$ end-to-end speedup while maintaining comparable image quality.The code is available at \url{https://github.com/hebutBryant/RegionCache}.

\end{abstract}

\input{new_sec1}

\input{sec2}

\input{sec3}

\input{sec4}
\input{sec5}

\input{sec6}

\bibliographystyle{named}
\bibliography{ijcai26}

\end{document}

%% file: new_sec1.tex
\section{Introduction}
Image generation is a core component of modern visual content applications~\cite{dhariwal2021diffusion,rombach2022high}. In practice, it is rarely a one-shot process; users typically refine results through multiple editing turns~\cite{nichol2021improved,miyake2025negative} until the output matches their intent. We refer to this setting as post-generation \textbf{multi-turn image editing}, where images are progressively refined by repeatedly re-running the diffusion process. Existing image editing methods adopt Diffusion Transformers (DiTs) due to their strong generation quality~\cite{labs2025flux1kontextflowmatching,esser2024scaling,chen2024pixart}. However, DiT inference is computationally expensive, and repeating it across multiple editing turns leads to substantial redundant computation overhead.



\begin{figure}[t]
    \centering

    \begin{subfigure}[t]{0.90\linewidth}
        \centering
        \includegraphics[
            width=\linewidth
        ]{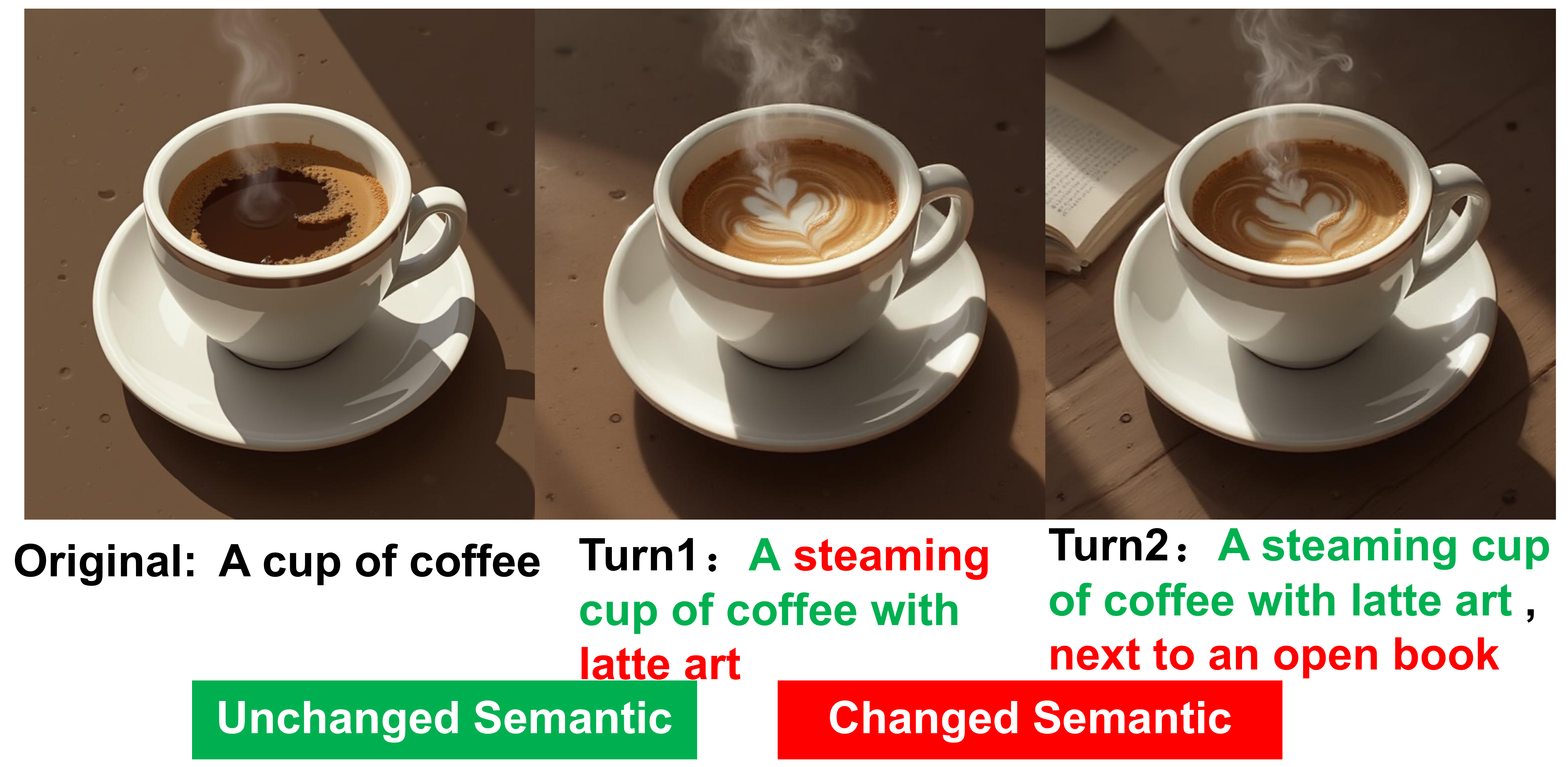}
        \vspace{-0.5em}
        \caption{Example of multi-turn image editing.}
        \label{fig:multiturn_a}
    \end{subfigure}

    \vspace{-0.2em}

    \begin{subfigure}[t]{0.90\linewidth}
        \centering
        \includegraphics[
            width=\linewidth
        ]{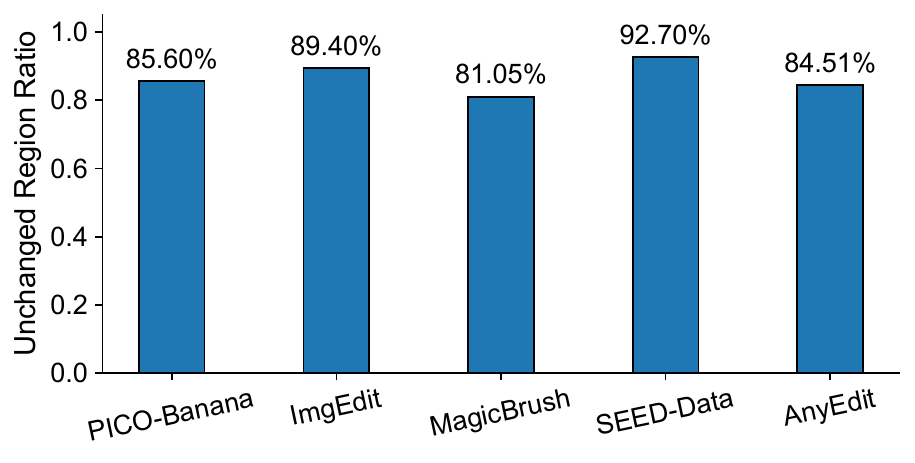}
        \vspace{-0.5em}
        \caption{Unchanged region ratio in image edit dataset.}
        \label{fig:pixel_b}
    \end{subfigure}

    \vspace{-0.7em}
    \caption{Multi-turn image editing illustration.}
    \vspace{-0.7em}
    \label{fig:multiturn_ab}
\end{figure}

Through an analysis of real-world multi-turn image editing datasets, we identify two key observations.
First, edits are typically spatially localized. In each turn, users usually modify a small, concentrated region of the image, while the remaining content is expected to stay unchanged (e.g., adjusting the coffee appearance while preserving the cup and background in Figure~\ref{fig:multiturn_a}. Across five real-world datasets~\cite{yu2025anyedit,ge2024seed,zhang2023magicbrush,ye2025imgedit,qian2025pico}, we quantify cross-turn image similarity using pixel-level overlap and find that consecutive turns share 81.05\%–92.70\% unchanged regions on average.
Second, edit instructions exhibit strong semantic persistence. Consecutive prompts largely repeat the same semantic content and differ only by a small incremental change. This semantic consistency naturally indicates which image content should be preserved across turns, while the prompt difference highlights the regions that require modification.


These observations suggest an opportunity to reduce redundant computation in multi-turn image editing by exploiting cross-turn semantic consistency. Since most of the image remains unchanged and most prompt semantics persist across turns, a large portion of the computation is repeatedly spent on the same content. In particular, semantics that are repeated in consecutive prompts typically correspond to unchanged image regions. By aligning these repeated semantics with their spatial support, we can reuse intermediate diffusion states from previous turns for those regions, avoiding full-image attention recomputation while focusing updates only on regions implied by the prompt delta.


Despite the intuitive opportunity for cross-turn computation reuse, existing DiT inference acceleration methods that exploit computation reuse are primarily designed for single-pass image generation~\cite{ma2024deepcache,zouaccelerating,liu2024faster}. These methods typically reduce inference cost by skipping denoising steps~\cite{ma2024deepcache,liu2024faster} or pruning attention~\cite{zouaccelerating,bolya2023token} within a single editing turn, and their reuse decisions are made solely based on intra-run characteristics. As a result, they do not directly address the dominant inefficiency in multi-turn editing, where similar content repeatedly appears across successive editing turns and redundancy arises across runs. Moreover, prior works~\cite{rombach2022high,zhou2025multi} have shown that aggressive approximation within a single inference run can degrade image quality, since high-quality editing often relies on dense attention and gradual refinement. These limitations motivate a more reliable reuse strategy tailored to multi-turn editing, which explicitly leverages cross-turn semantic consistency.



In this work, we propose \textbf{RegionCache}, an end-to-end framework for efficient multi-turn image editing that combines prompt-aligned region caching with a semantic-aware region reusing. Given a new editing prompt over a generated image, RegionCache identifies regions whose diffusion states can be safely reused by aligning shared semantics across consecutive prompts and localizing their spatial support via cross-attention maps. During inference, RegionCache reuses cached diffusion states for unchanged regions and recomputes only the regions affected by the prompt update. 
To balance efficiency and quality, RegionCache further determines the {reuse depth} based on prompt semantic similarity and editing context: semantically consistent edits permit reuse for more diffusion steps, while larger semantic changes reduce the reuse depth and fall back to reusing only early steps (e.g., \textit{cat} $\rightarrow$ \textit{dog}). This design avoids unnecessary recomputation while maintaining high editing fidelity in interactive, text-driven workflows.



The main contributions of this paper are summarized as follows:

\begin{itemize}
    \item We identify cross-turn {region-level semantic stability} as a key source of redundant computation in multi-turn image editing, and propose a prompt-aligned region caching mechanism that localizes reusable regions via cross-attention.
    \item We propose a {semantic-aware region reusing policy} that adaptively determines the safe reuse horizon based on prompt semantic similarity and editing context, enabling aggressive reuse for semantically consistent edits and conservative reuse for larger semantic changes.
    \item We design \textbf{RegionCache}, an end-to-end framework for DiT-based multi-turn image editing that integrates region-level caching with adaptive step reuse, achieving significant inference speedups while preserving high editing fidelity.
\end{itemize}

%% file: sec2.tex
\section{Related Work}

\subsection{Diffusion Transformer Inference Acceleration Methods }
Diffusion Transformers (DiTs)~\cite{peebles2023scalable,chen2024pixart} have become a popular backbone for high-quality image generation by modeling latent patches with global self-attention. While effective, DiT inference is computationally expensive due to the iterative denoising process and the quadratic complexity of self-attention.
This has motivated a large body of work on accelerating diffusion inference.

Many acceleration methods for diffusion models exploit the observation that intermediate representations evolve smoothly across denoising steps, allowing partial reuse of computation~\cite{zhang2025blockdance,chen2024delta,yu2025ab,liu2024faster}. For example, DeepCache~\cite{ma2024deepcache} identifies diffusion steps where feature changes are limited and reuses previously computed representations, thereby skipping redundant computation. $\Delta$-DiT~\cite{chen2024delta} introduces a $\Delta$-Cache that stores feature deviations instead of feature maps, enabling block skipping in DiT for efficient acceleration. BlockDance~\cite{zhang2025blockdance} accelerates diffusion inference by reusing structural features across consecutive steps and skipping the computation of shallow Transformer blocks within each denoising step.

Another line of work focuses on reducing the cost of attention computation, which dominates the runtime of Diffusion Transformers. These methods dynamically reduce attention computation by identifying or predicting which tokens contribute less to generation~\cite{zhang2025sla,yuan2024ditfastattn,bolya2023token}. 
For example, TGATE~\cite{liu2024faster} observes that the importance of attention computation varies across diffusion stages, and reduces inference cost by selectively reusing attention outputs at stages where their contribution becomes less critical. 
RAS~\cite{liu2025region} and ToCa~\cite{zouaccelerating}, which predict redundant tokens and bypass their attention computation.

Overall, existing DiT inference acceleration methods are primarily designed for single-pass generation. Their reuse decisions rely on intra-run characteristics, such as step-wise similarity or token importance, and do not capture redundancy that arises across successive editing turns. When applied to multi-turn editing, these methods either fail to reduce cross-turn redundancy or introduce noticeable quality degradation due to aggressive approximation.

\subsection{Multi-turn Image Editing}
Multi-turn image editing~\cite{joseph2024iterative,zhou2025multi} supports iterative refinement, where each turn modifies an existing image to better match user intent.
The process starts from a given image—either real or previously generated—and applies a sequence of edits while preserving appropriate aspects of the image across turns.
This reflects common user workflows, where images are refined over multiple turns until a satisfactory result is obtained.

\paragraph{Training-based methods.}
Training-based methods ~\cite{esser2024scaling,wang2023stylediffusion,huang2024diffstyler} explicitly optimize diffusion models for editing by introducing additional supervision, fine-tuning, or architectural extensions (e.g., paired editing trajectories, region-aware annotations, or instruction-tuning on large synthetic datasets). Such designs often yield strong editability and robustness, but require extra data collection and training/optimization cost, making them less flexible for interactive multi-turn use.



\paragraph{Training-free methods.}
Several training-free methods~\cite{tumanyan2023plug,couairondiffedit,cao2023masactrl} address multi-turn image editing by modifying the inference behavior of pretrained diffusion models. These approaches mainly focus on improving {editing controllability and visual consistency}, for example by constraining attention or injecting reference information from previous generations. Prompt-to-Prompt (P2P)~\cite{hertzprompt} steers cross-attention maps to localize edits while preserving unrelated content, whereas StableFlow~\cite{avrahami2025stable} maintains consistency by injecting reference features during inference. Region-Aware Generation (RAG)~\cite{chen2024region} further introduces region-aware constraints to guide localized generation.

However, these methods primarily operate at the {output or attention guidance level} and do not change the underlying inference execution. In practice, the diffusion model still performs full-image denoising and attention computation at every step, even when the edit affects only a small region. As a result, while these approaches improve editing quality and stability, they do not reduce the computational cost of multi-turn editing, and repeated full-image inference remains the dominant source of latency.

%% file: sec3.tex
\section{RegionCache}




\begin{figure}[t]
    \centering
    \includegraphics[width=\linewidth]{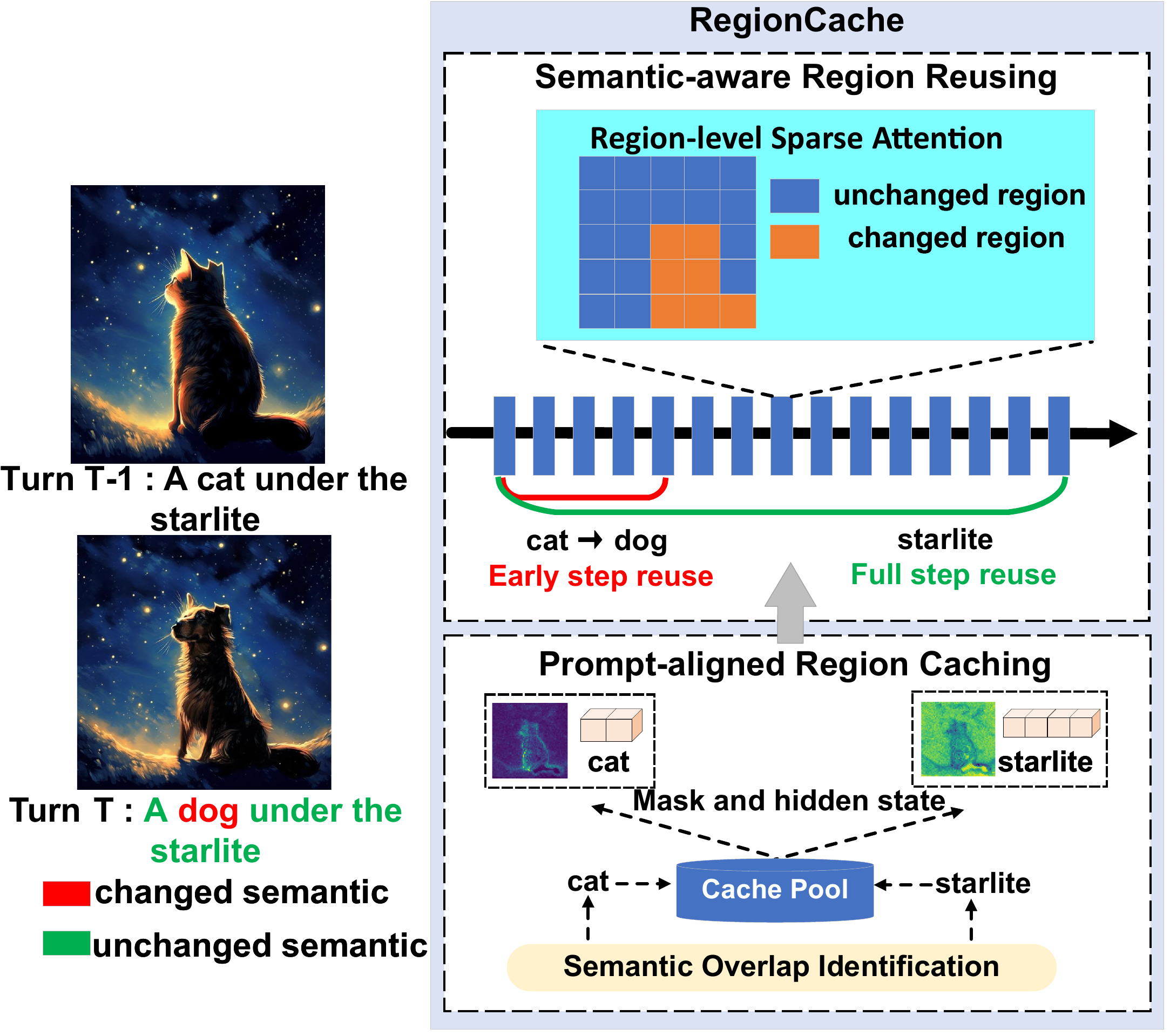}
    \vspace{-0.4em}
    \caption{An overview of RegionCache.}
    \vspace{-0.6em}
    \label{fig:overview}
\end{figure}

\subsection{Overview}
In this section, we present the design principles and key techniques of RegionCache for accelerating multi-turn image editing with Diffusion Transformers. RegionCache exploits both region-level and timestep-level reuse to reduce redundant computation across editing turns.


Figure \ref{fig:overview} shows an overview.
Specifically, RegionCache consists of two core components. First, a {prompt-aligned region caching} mechanism reuses intermediate diffusion states for regions whose semantics persist across consecutive prompts, while recomputing only regions implicated by the new edit. Second, a {semantic-aware region reuse} strategy decides {how far} each cached region can be reused along the denoising trajectory. It leverages prompt similarity and editing context to choose a reuse depth: semantically consistent edits allow reuse for more steps, whereas larger semantic changes restrict reuse to early steps and progressively refresh the region when needed. Together, these components enable efficient multi-turn editing while preserving generation quality.

\begin{figure}[t]
    \centering
    \includegraphics[width=\linewidth]{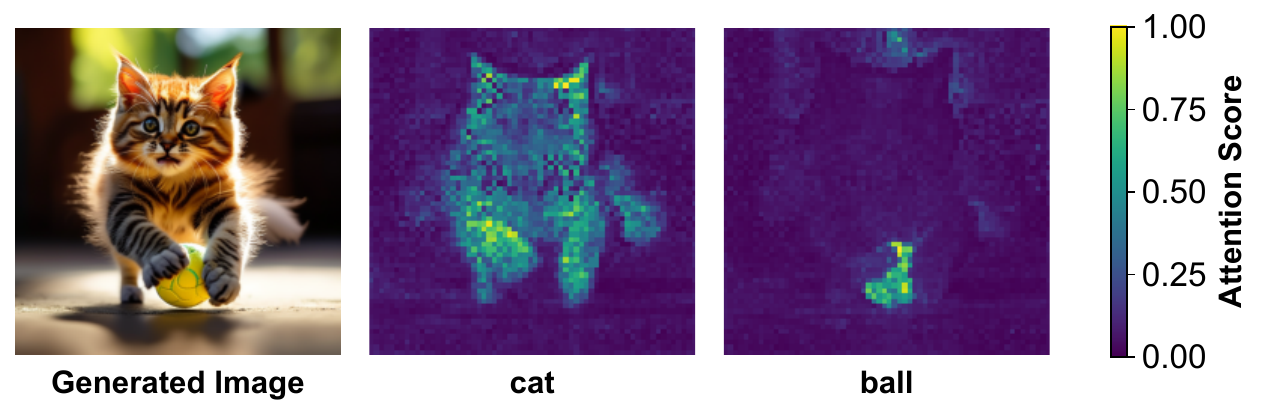}
    \caption{
    Token-level cross-attention visualization.
    }
    \label{fig:token_attention}
\end{figure}

\subsection{Prompt-aligned Region Caching  }

RegionCache localizes reusable content by exploiting semantic consistency across consecutive editing prompts. Given two prompts from successive turns, we first identify which semantics persist and which are newly introduced by comparing their tokens. Persistent semantics indicate image content that should be preserved and reused, while newly introduced or modified semantics specify where the edit should take effect. To reuse diffusion states accordingly, these semantic differences must be translated into spatial regions in the image.

Diffusion Transformers provide a natural mechanism for this translation through cross-attention. During generation, cross-attention explicitly links text tokens to the spatial locations they influence in the latent image. As illustrated in Fig.~3, tokens describing concrete objects (e.g., \emph{cat} or \emph{ball}) typically attend to compact and contiguous regions, with negligible attention on unrelated areas. RegionCache leverages this property to align prompt semantics with image regions, enabling region-level caching of diffusion states.

Concretely, RegionCache localizes edit regions using cross-attention maps produced during diffusion inference. Let $A_t \in \mathbb{R}^{N \times M}$ denote the cross-attention matrix at denoising step $t$, where $N$ and $M$ are the numbers of image and text tokens. For each newly introduced or modified token $j$, we aggregate its attention scores across heads and layers to obtain a spatial attention vector $a_j \in \mathbb{R}^{N}$. Image tokens with salient attention responses are considered affected by the edit. Formally, the edit region is defined as
\[
\mathcal{R}_{\text{edit}} = \left\{ i \mid a_j(i) \ge \theta_j \right\},
\]
where $\theta_j$ is an attention-based threshold derived from $a_j$. Regions outside $\mathcal{R}_{\text{edit}}$ correspond to persistent semantics and are treated as reusable; their cached diffusion states are preserved for subsequent editing turns.

\paragraph{Hidden-State-Oriented Caching.}
After localizing the regions affected by an edit, RegionCache avoids redundant computation by reusing intermediate {hidden states} of the remaining regions across editing turns. In Diffusion Transformers, hidden states are the internal representations processed at each denoising step: they are the direct inputs and outputs of self-attention and MLP layers, and fully determine the computation performed at that step. Consequently, hidden states dominate the inference cost in DiT-based editing.

Caching hidden states therefore provides a direct and effective way to reduce computation. By reusing the hidden states of semantically stable regions, RegionCache can skip repeated self-attention and MLP computation for those regions while preserving their denoising behavior. This enables reuse at the exact level where most computation occurs, without approximating or altering the model’s execution.

To support multi-turn editing, RegionCache maintains a cache pool of hidden states along the diffusion trajectory. During a new editing turn, cached hidden states corresponding to unchanged regions are retrieved and reused, while hidden states of edit-affected regions are recomputed and updated. This design enables efficient cross-turn reuse with minimal impact on generation quality.

\begin{figure}[t]
    \centering
    \includegraphics[width=\linewidth]{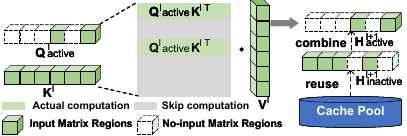}
    \vspace{-0.5em}
    \caption{Region-level sparse attention in the transformer block.}
    \vspace{-0.7em}
    \label{fig:compute}
\end{figure}

\subsection{Semantic-aware Region Reusing}

\paragraph{Region-Level Sparse Attention.}
To reuse cached hidden states while avoiding redundant attention computation, RegionCache introduces {region-level sparse attention}. The key idea is to perform self-attention only on regions affected by the current edit, while directly reusing cached hidden states for all other regions. This design preserves the standard DiT execution semantics for edited regions, while eliminating unnecessary computation on unchanged content.

As illustrated in Figure.~\ref{fig:compute}, region-level sparse attention is implemented inside each Transformer block. Based on the region identification results, image tokens are partitioned into {active regions} that require updating and {inactive regions} whose hidden states are reused. For a given layer $l$, only the hidden states of active regions are instantiated as queries, denoted as $Q^l_{\text{active}}$, while the key and value matrices $K^l$ and $V^l$ are shared across all regions and remain unchanged.

During attention computation, only $Q^l_{\text{active}}$ participates in the matrix multiplication with $K^l$, producing attention outputs for the active regions. Hidden states corresponding to inactive regions are not involved in attention at all; instead, their updated representations $H^{l+1}_{\text{inactive}}$ are directly retrieved from the cache. The outputs from attention, $H^{l+1}_{\text{active}}$, are then merged with the cached inactive hidden states to form the complete hidden representation $H^{l+1}$, which is passed to the next Transformer block following the standard DiT inference pipeline.

\begin{figure}[t]
    \centering
    \includegraphics[width=\linewidth]{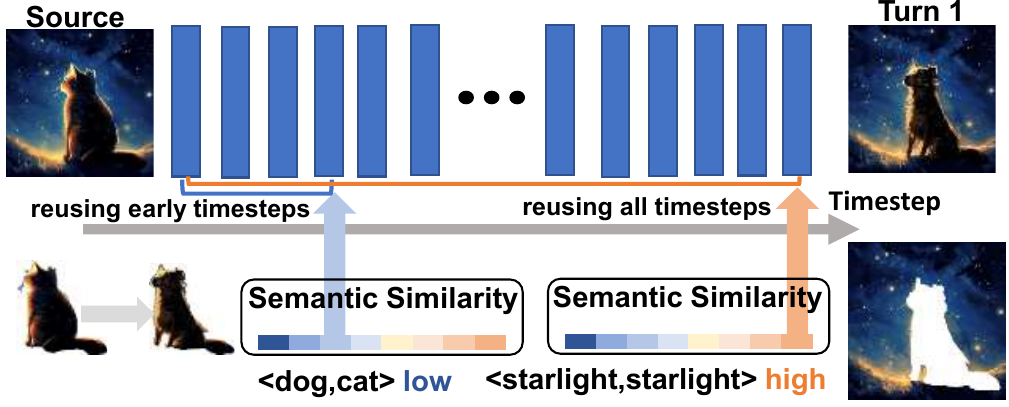}
    \vspace{-0.5em}
    \caption{Adaptive timestep reuse driven by semantic similarity.}
    \vspace{-0.7em}
    \label{fig:schedular}
\end{figure}

\paragraph{Semantic-aware Adaptive Reuse Scheduling.}
Diffusion inference progresses through a sequence of denoising steps that play different roles in image generation. Prior studies~\cite{chen2024delta,zhang2025blockdance,agarwal2024approximate,you2025layer} have shown that early steps primarily establish coarse structure, while later steps focus on refining object-specific details and appearance. This property implies that {semantic changes do not affect all diffusion steps equally}.

RegionCache leverages this observation to enable adaptive reuse along the diffusion trajectory. Even when a region undergoes a semantic edit, such as changing \emph{cat} to \emph{dog}, the impact of this change typically emerges at later denoising steps where fine-grained semantics are resolved. The hidden states at early steps often remain compatible and can be safely reused, whereas recomputation becomes necessary only after a certain point. In contrast, for regions whose semantics remain unchanged across editing turns, cached hidden states can be reused throughout the entire diffusion process without affecting generation quality, as illustrated in Figure~\ref{fig:schedular}.

Based on this insight, RegionCache adopts a {semantic-aware adaptive reuse scheduling} strategy that determines how long cached hidden states can be reused for each region. Specifically, for each edit-affected region, we associate the corresponding text tokens before and after the edit, denoted as $\tau^{-}$ and $\tau^{+}$. We quantify their semantic change using cosine similarity:
\begin{equation}
s = \mathrm{sim}\!\left(\mathbf{e}(\tau^{-}),\, \mathbf{e}(\tau^{+})\right),
\end{equation}
where $\mathbf{e}(\cdot)$ denotes the token embedding.

If the similarity score exceeds a threshold $\tau_s$, the region is considered semantically consistent across turns, and its cached hidden states are reused throughout the entire diffusion process. Otherwise, the region is treated as undergoing a semantic change, and reuse is restricted to early denoising steps only. Following prior observations that early diffusion steps mainly capture coarse structure\cite{chen2024delta,zhang2025blockdance}, we reuse cached hidden states for the first $\rho T$ denoising steps and recompute the region thereafter, where $T$ denotes the total number of denoising steps. Based on established practice in prior work, we set the reuse ratio to a fixed value $\rho = 0.2$ in our implementation.


Formally, the reuse cutoff is defined as
\begin{equation}
T_{\text{reuse}} =
\begin{cases}
T, & s \ge \tau_s, \\
\lfloor \rho T \rfloor, & s < \tau_s .
\end{cases}
\end{equation}

This design allows RegionCache to aggressively reuse computation for semantically stable regions, while conservatively limiting reuse for larger semantic edits, achieving a favorable balance between efficiency and editing fidelity.

\subsection{Implementation}
\paragraph{Kernel Fusing.}
Region Sparse Attention updates keys and values only for active tokens, while reusing cached ones for the rest. A naive implementation would require extra scatter operations, introducing additional memory access and kernel overhead. To avoid this, we fuse the scatter operation into the linear projection GeMM kernel, directly writing active keys and values to their target positions. This fusion eliminates redundant kernels and ensures that sparse attention yields a practical speedup.
\paragraph{Pipelined Execution.}
Cached hidden states occupy substantial memory and are therefore stored in host (CPU) memory. To reduce the overhead of transferring cached features during inference, RegionCache overlaps cache loading with computation using a pipelined execution strategy~\cite{harris2012overlap}. While active regions are being processed in one layer, cached features for the next layer are prefetched, effectively hiding data transfer cost and improving overall efficiency.

%% file: sec4.tex
\section{Experiment}

\subsection{Experiment Setup}
\paragraph{Benchmarks and Task Setting}
We evaluate RegionCache under a unified multi-turn image editing setting based on the PICO-Banana~\cite{qian2025pico} and MagicBrush~\cite{zhang2023magicbrush} datasets. Each sample consists of a sequence of prompt-driven edits, where each turn incrementally refines the result from the previous turn, reflecting realistic interactive editing scenarios. Unless otherwise specified, all methods are evaluated on identical editing sequences and region specifications.

\paragraph{Evaluation Metrics}
We evaluate both efficiency and editing quality. Efficiency is measured by the total wall-clock latency required to complete an entire editing sequence, with attention latency additionally reported to analyze reductions in the dominant computational cost.

For generating quality under inference acceleration methods, we report FID~\cite{zhang2018unreasonable}, CLIP~\cite{radford2021learning}, PickScore~\cite{kirstain2023pick}, and Image Relevance (IR) to measure overall image quality and text--image semantic alignment.

For multi-turn editing performance, we adopt turn-aware CLIP-based metrics. Specifically, CLIP$_{\text{txt}}$ measures prompt alignment at each editing turn, CLIP$_{\text{img}}$ evaluates cross-turn visual consistency between consecutive outputs, and CLIP$_{\text{dir}}$ assesses whether edits follow the intended semantic direction.

\paragraph{Single-turn acceleration baselines.}
We compare RegionCache with representative inference acceleration and caching methods originally designed for single-turn generation, including
ToCa~\cite{zouaccelerating}, DeepCache~\cite{ma2024deepcache}, TGATE~\cite{liu2024faster}, and step-reduced PixArt-LCM~\cite{chenpixart},
all implemented on the same PixArt backbone for fair comparison. To evaluate these methods in a multi-turn setting, we adopt a unified editing protocol in which, at each turn, edits are applied to specified regions while unedited regions are preserved
by constraining these methods to reuse and maintain the corresponding latent representations.

\paragraph{Multi-turn image editing baselines.}
We further compare RegionCache with representative training-free multi-turn image editing frameworks, including StableFlow~\cite{avrahami2025stable}, RAG (Region-Aware Generation)~\cite{chen2024region},
and Prompt-to-Prompt (P-2-P)~\cite{hertzprompt}, under the same multi-turn editing protocol. This comparison focuses on evaluating efficiency and editing quality under realistic interactive editing scenarios.
All methods are evaluated using the pretrained PixArt-$\alpha$ Diffusion Transformer under identical inference settings to ensure fair comparison. For caching-based acceleration baselines originally designed for single-turn generation, we adapt their strategies to the multi-turn setting by applying caching independently at each editing turn. All experiments are conducted on a single NVIDIA RTX L20 GPU, and results are reported over 30,000 randomly sampled multi-turn editing sequences unless otherwise specified.

\begin{table}[t]
    \centering
    \small
    \setlength{\tabcolsep}{3pt}
    \renewcommand{\arraystretch}{1.05}
    \begin{tabular*}{\columnwidth}{@{\extracolsep{\fill}}l c c c c}
        \hline
        \textbf{Method}
        & Latency (s)$\downarrow$
        & CLIP$_{txt}\uparrow$
        & CLIP$_{img}\uparrow$
        & CLIP$_{dir}\uparrow$ \\
        \hline
        Stable Flow
            & 17.46
            & 0.3097
            & 0.8853
            & 0.072 \\
        P-2-P
            & 17.19
            & 0.2779
            & 0.9604
            & 0.103 \\
        RAG
            & 32.30
            & 0.3159
            & \textbf{0.9678}
            & \textbf{0.120} \\
        \textbf{RegionCache}
            & \textbf{11.11}
            & \textbf{0.3204}
            & 0.9233
            & 0.101 \\
        \hline
    \end{tabular*}
    \vspace{-0.5em}
    \caption{Quantitative comparison of multi-turn image editing methods on the MagicBrush dataset, where $\uparrow$ indicates higher is better and $\downarrow$ indicates lower is better.}
    \vspace{-0.7em}
    \label{tab:clip_metrics}
\end{table}

\begin{figure}[t]
    \centering
    \includegraphics[width=\columnwidth]{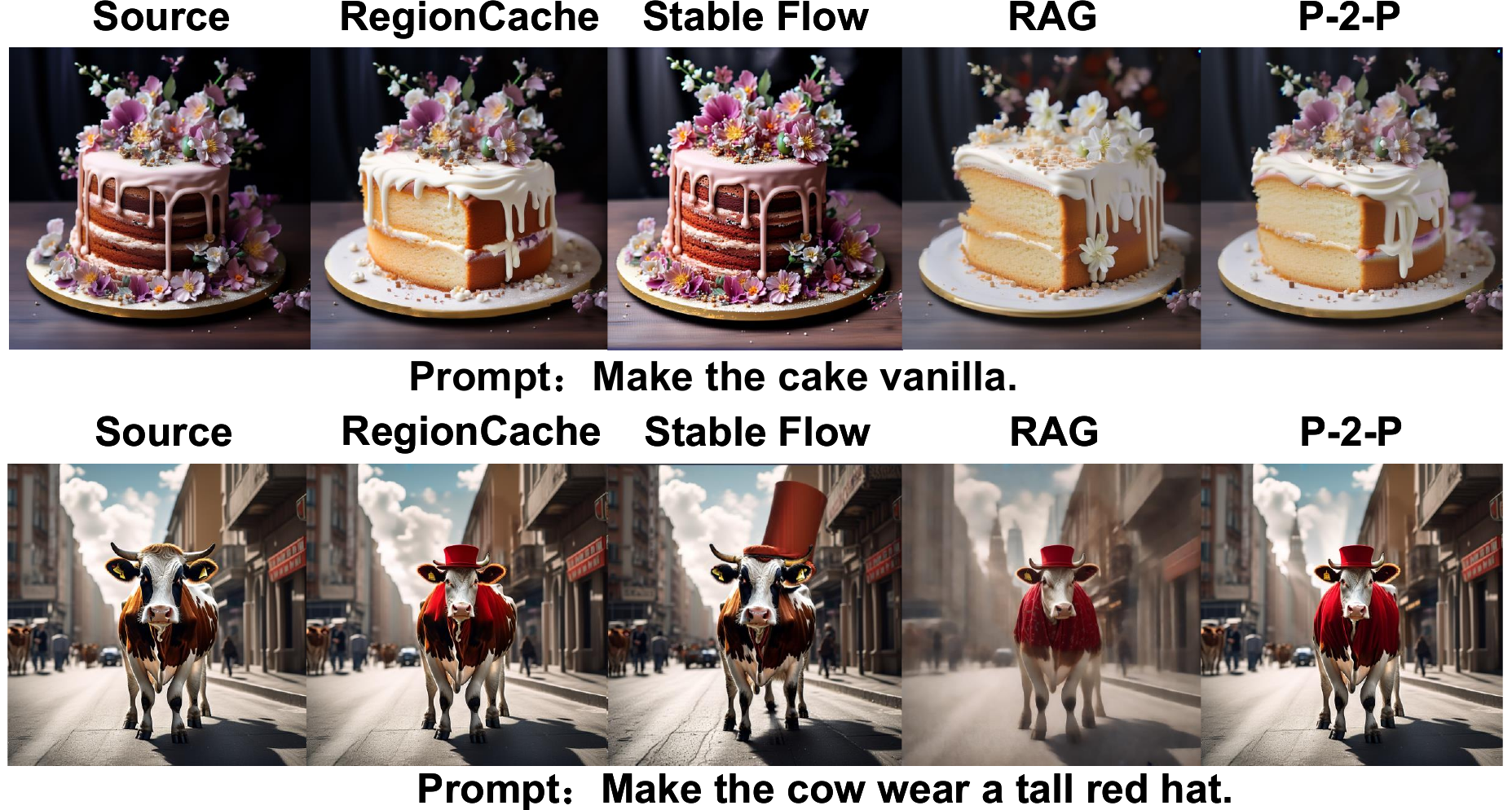}
    \vspace{-0.5em}
    \caption{Case study of multi-turn image editing comparing RegionCache with existing multi-turn editing frameworks.}
    \vspace{-0.7em}
    \label{fig:example_edit2}
\end{figure}

\subsection{Comparison with Multi-turn Image Editing Frameworks}
As shown in Table~\ref{tab:clip_metrics}, RegionCache achieves the lowest overall latency, completing a full multi-turn editing sequence in 11.11~s. This corresponds to a $1.6\times$ speedup over Stable Flow (17.46~s) and P-2-P (17.19~s),
and a $2.9\times$ speedup over RAG (32.30~s). For Stable Flow and P-2-P, the additional runtime mainly stems from recomputing full-image self-attention at every editing turn, which introduces substantial redundant computation
when large regions remain unchanged.
On the other hand, RAG incurs higher latency due to the overhead introduced by explicit
region-aware control.

RegionCache also maintains competitive editing quality~\ref{fig:example_edit2}.
It achieves the highest CLIP$_{\text{txt}}$ score (0.3204), indicating superior alignment with textual prompts across editing turns. This improvement stems from RegionCache’s region-level reuse mechanism, which preserves semantically consistent representations for unchanged regions, thereby reducing unintended semantic drift during successive edits.
While its CLIP$_{\text{img}}$ score is slightly lower than that of P-2-P and RAG, the difference remains moderate, suggesting that RegionCache preserves cross-turn visual consistency while allowing necessary local modifications to accommodate updated prompts. In addition, RegionCache attains a comparable CLIP$_{\text{dir}}$ score, reflecting stable and coherent editing directions, as semantic changes are explicitly constrained to edited regions.

Overall, these results indicate that RegionCache strikes a favorable balance between efficiency and editing quality.

\subsection{Comparison with Single-turn Acceleration Methods}
We also compare RegionCache with existing diffusion acceleration methods under a unified multi-turn setting.
As shown in Table~\ref{tab:cache_comparison}, RegionCache achieves a favorable balance between efficiency and image quality, outperforming the majority of compared acceleration methods in both inference speed and editing fidelity.
RegionCache reduces the total inference latency to 2.53s, achieving a $1.57\times$ speedup over the full PixArt-$\alpha$ baseline (3.98s), while lowering the cumulative self-attention computation time across all denoising steps to 0.55~s, corresponding to a $2.89\times$ reduction.   
Specifically, RegionCache is faster than ToCa by 0.02~s in total inference time and reduces the cumulative self-attention cost by 0.25~s (0.55~s vs.\ 0.80~s), as region-level reuse enables a higher effective degree of attention sparsification across editing turns.
Compared with TGATE, RegionCache further achieves a 0.15~s reduction in total inference time (2.53~s vs.\ 2.68~s) and lowers the attention cost by 0.34~s (0.55~s vs.\ 0.89~s), despite TGATE incorporating partial reuse of attention computation results, as TGATE still retains full self-attention at selected denoising steps.
DeepCache attains a higher total inference speedup of $1.81\times$ by aggressively skipping diffusion steps, and therefore surpasses RegionCache in raw speedup.
However, this aggressive step skipping comes at a clear cost in editing quality: DeepCache yields substantially worse FID and IR than RegionCache.

\begin{figure*}[t]
    \centering
    \includegraphics[width=0.55\linewidth]{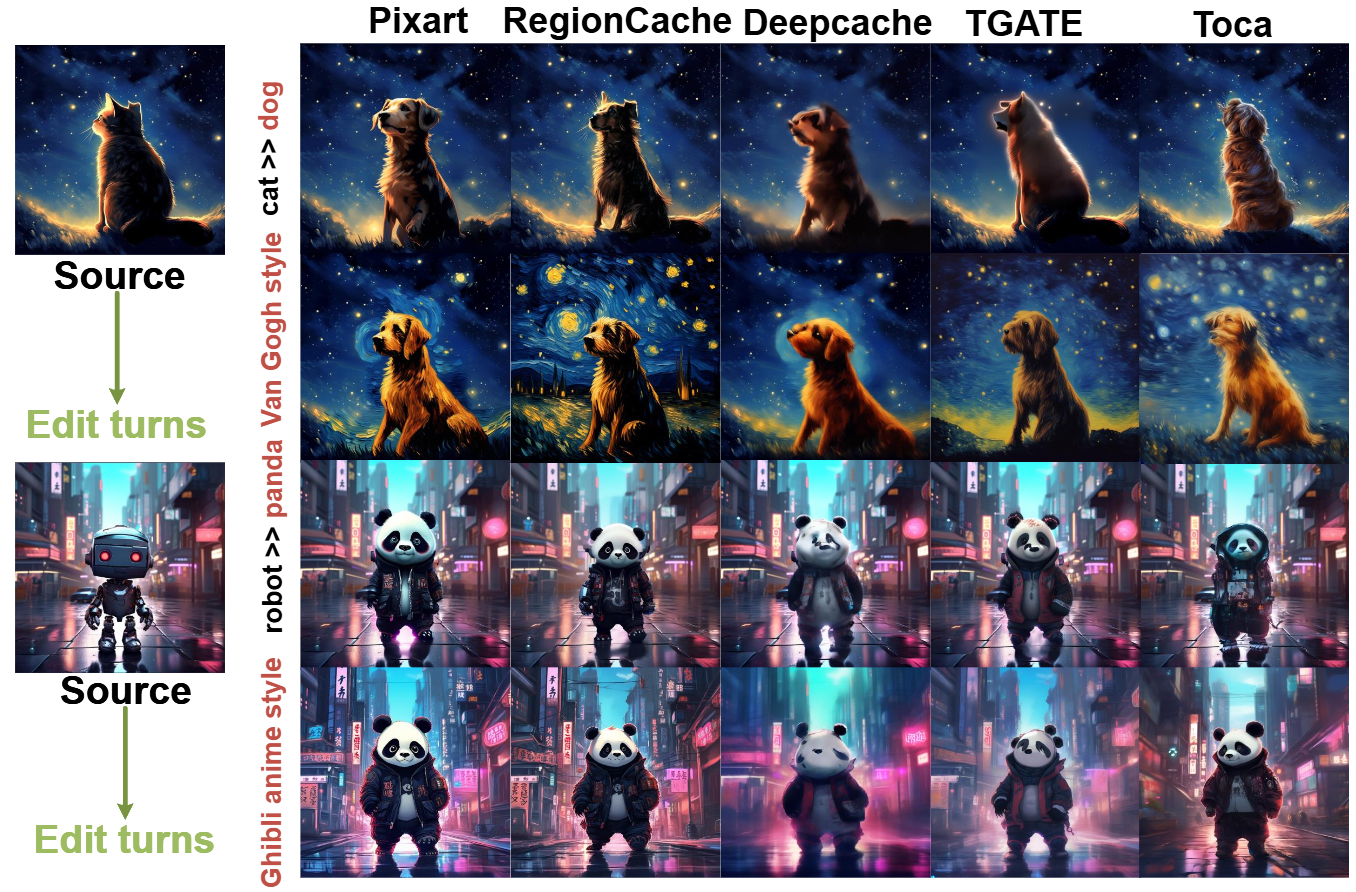} 
    \vspace{-0.5em}
    \caption{Case study of multi-turn image editing comparing RegionCache with existing diffusion acceleration methods.}
    \vspace{-0.7em}
    \label{fig:example_inpaint}
\end{figure*}

RegionCache achieves the best overall image quality because it reuses a more stable caching target in multi-turn editing: hidden states from regions whose semantics persist across turns.
As shown in Table~\ref{tab:cache_comparison}, RegionCache attains the lowest FID score and the highest CLIP, IR, and PICK metrics, indicating superior semantic alignment and fine-grained editing fidelity.
This advantage is further illustrated in Figure~\ref{fig:example_inpaint}, where RegionCache produces more coherent local edits and preserves better structural consistency across multiple editing turns.

\begin{table}[t]
\centering
\small
\setlength{\tabcolsep}{1.8pt}
\renewcommand{\arraystretch}{1.12}

\begin{tabular}{p{2.15cm} cc cccc}
\hline
& \multicolumn{2}{c}{\textbf{Inference Time}}
& \multicolumn{4}{c}{\textbf{Image Quality}} \\
\cline{2-3}\cline{4-7}

\textbf{Method}
& \textbf{Attn. (s)}$\downarrow$
& \textbf{Total (s)}$\downarrow$
& \textbf{FID}$\downarrow$
& \textbf{CLIP}$\uparrow$
& \textbf{IR}$\uparrow$
& \textbf{PICK}$\uparrow$ \\
\hline

PixArt-$\alpha$ (Full compute)
& 1.58
& 3.98
& \textbf{30.30}
& \textbf{0.311}
& 0.747
& \textbf{22.41} \\

ToCa ($N{=}3$, $R{=}60\%$)
& 0.80
& 2.55
& 36.65
& 0.303
& 0.172
& 21.02 \\

DeepCache ($N{=}3$)
& 0.73
& \textbf{2.20}
& 39.65
& 0.307
& 0.186
& 21.14 \\

TGATE ($m{=}10$)
& 0.89
& 2.68
& 31.27
& 0.211
& 0.678
& 21.99 \\

PixArt-LCM (8)
& 0.44
& 1.03
& 33.44
& 0.307
& 0.485
& 21.98 \\

RegionCache
& \textbf{0.55}
& 2.53
& \textbf{30.98}
& \textbf{0.311}
& \textbf{0.776}
& \textbf{22.36} \\
\hline
\end{tabular}

\vspace{-0.5em}
\caption{
Comparison with existing multi-turn image editing methods on the PICO-Banana dataset.
$\uparrow$ indicates higher is better, while $\downarrow$ indicates lower is better.
}
\vspace{-0.7em}

\label{tab:cache_comparison}
\end{table}

\begin{figure}[t]
    \centering
    \includegraphics[
        width=0.95\linewidth
    ]{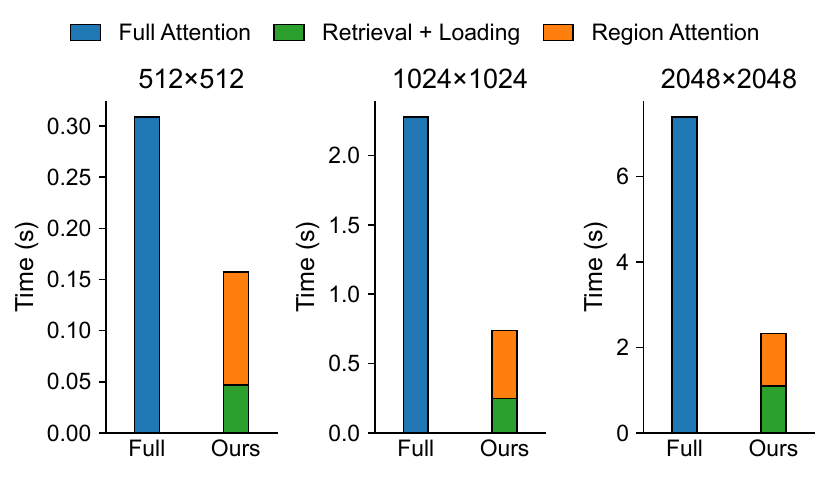}
    \vspace{-0.5em}
    \caption{Resolution-wise inference overhead analysis.}
    \vspace{-0.7em}
    \label{fig:resolution_overhead}
\end{figure}

\begin{figure*}[t]
    \centering
    \includegraphics[width=0.95\textwidth]{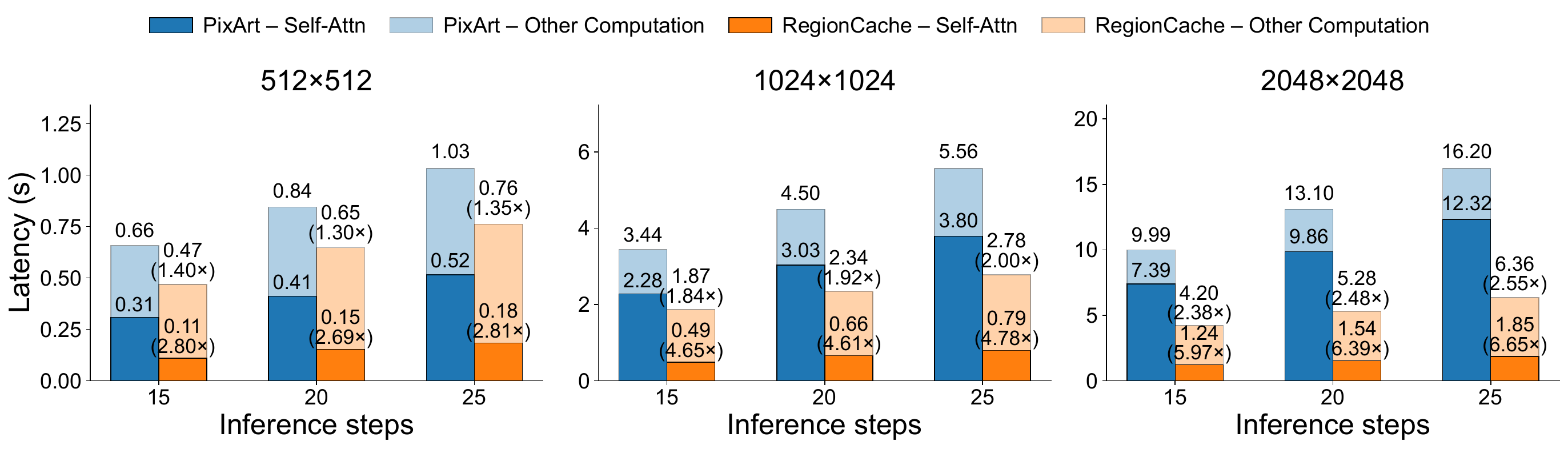}
    \vspace{-0.5em}
    \caption{Latency breakdown of self-attention and other computation across different inference steps and resolutions.}
    \vspace{-0.7em}
    \label{fig:latency_breakdown}
\end{figure*}

\subsection{Runtime Breakdown Analysis}
In a single inference pass, RegionCache introduces two additional sources of overhead: the cost of retrieving reusable cached regions and the cost of loading the corresponding cached hidden states from host memory into device memory.

Figure~\ref{fig:resolution_overhead} compares these overheads with the reduction in attention computation across different image resolutions. As resolution increases, the cost of full attention grows rapidly, from around 0.3~s at $512\times512$ to over 2~s at $1024\times1024$, and exceeds 6~s at $2048\times2048$. In contrast, the combined retrieval and loading overhead remains small, on the order of 0.05~s at $512\times512$ and below 0.3~s even at $2048\times2048$. Across all resolutions, RegionCache reduces the attention computation by a large margin, with region-level attention accounting for only a fraction of the full attention cost. As a result, the reduction in attention computation consistently outweighs the introduced overhead, leading to a net decrease in inference time. These results indicate that RegionCache effectively targets the dominant computation
while introducing only minor additional cost.

\subsection{Sensitive Study}
\paragraph{Performance with varying steps}
For a fixed image resolution, increasing the number of diffusion steps primarily scales the absolute runtime of all computation components in a nearly linear manner. As shown in Figure ~\ref{fig:latency_breakdown}, at $512\times512$ resolution, the self-attention latency of PixArt increases from 0.31~s at 15 steps to 0.52~s at 25 steps, while RegionCache increases from 0.11~s to 0.18~s over the same range. Despite this increase in absolute cost, the relative reduction in self-attention introduced by RegionCache remains stable at approximately $2.8\times$--$2.9\times$ across 15, 20, and 25 steps. A similar trend is observed at higher resolutions. This shows that varying the number of diffusion steps has a limited impact on the relative speedup achieved by RegionCache.

\paragraph{Performance with varying image resolution}
Image resolution has a pronounced impact on the effectiveness of RegionCache.
In Figure ~\ref{fig:latency_breakdown}, as resolution increases, self-attention rapidly becomes the dominant component of inference cost. At 15 diffusion steps, the self-attention latency of PixArt increases from 0.31~s at $512\times512$ to 7.39~s at $2048\times2048$, accounting for the majority of total runtime.
Although RegionCache applies a similar relative reduction to self-attention across resolutions, the absolute benefit grows substantially with image size. Specifically, the attention speedup increases from approximately $2.7\times$ at $512\times512$
to $4.7\times$ at $1024\times1024$, and exceeds $6\times$ at $2048\times2048$.

This amplification stems from the increased dominance of self-attention at larger sequence lengths. As self-attention occupies a larger portion of the total computation, the same relative sparsification ratio yields a greater reduction in end-to-end latency.

%% file: sec5.tex
\section{Conclusion}
We presented \textbf{RegionCache}, a region-level reuse framework for efficient multi-turn image editing with Diffusion Transformers. RegionCache identifies which parts of an image can be reused across editing turns by aligning repeated prompt semantics to image regions via cross-attention, and caches hidden states at the level where attention computation dominates. During editing, it selectively reuses cached hidden states for semantically stable regions and recomputes only the regions affected by new edits, with an adaptive reuse schedule that controls how long reuse remains valid along the diffusion trajectory.
Extensive experiments demonstrate that RegionCache significantly reduces redundant attention computation and achieves consistent end-to-end speedups while preserving editing quality.

\section*{Acknowledgments}
 This work was supported by the National Natural Science Foundation of China (62461146205, U2241212, 625B2041), and the Distinguished Youth Foundation of Liaoning Province (2024021148-JH3/501).